\documentclass[sigconf]{acmart}

\usepackage{booktabs} 

\setcopyright{rightsretained}

\acmConference[ICVGIP'25]{Indian Conference on Computer Vision, Graphics and Image Processing}{December 2025}{Mandi, India}
\acmYear{2025}
\copyrightyear{2025}

\acmPrice{15.00}

\begin{document}
\title{TinyViT: Field Deployable Transformer Pipeline for Solar Panel Surface Fault and Severity Screening}
\titlenote{Produces the permission block, and
  copyright information}

\author{Ishwaryah Pandiarajan}
\affiliation{%
  \institution{Thiagarajar College of Engineering}
  \streetaddress{XYZ}
  \city{Madurai}
  \state{Tamil Nadu}
  \country{India}
  \postcode{}}
  \email{ishwaryahp@student.tce.edu}
  \author{Mohamed Mansoor Roomi Sindha}
\affiliation{%
  \institution{Thiagarajar College of Engineering}
  \streetaddress{XYZ}
  \city{Madurai}
  \state{Tamil Nadu}
  \country{India}
  \postcode{}}
  \email{smmroomi@tce.edu}
  \author{Uma Maheswari Pandyan}
\affiliation{%
  \institution{Velammal College of Engineering and Technology}
  \streetaddress{XYZ}
  \city{Madurai}
  \state{Tamil Nadu}
  \country{India}
  \postcode{}}
  \email{umamahes.p@gmail.com}
  \author{Sharafia N}
\affiliation{%
  \institution{Kumaraguru College of Technology}
  \streetaddress{XYZ}
  \city{Coimbatore}
  \state{Tamil Nadu}
  \country{India}
  \postcode{}}
  \email{sharafiaahamed997@gmail.com}

\renewcommand{\shortauthors}{}

\begin{abstract}
Sustained operation of solar photovoltaic assets hinges on accurate detection and prioritization of surface faults across vast, geographically distributed modules. While multi modal imaging strategies are popular, they introduce logistical and economic barriers for routine farmlevel deployment. This work demonstrates that deep learning and classical machine learning may be judiciously combined to achieve robust surface anomaly categorization and severity estimation from planar visible band imagery alone. We introduce TinyViT which is a compact pipeline integrating Transformer based segmentation, spectral-spatial feature engineering, and ensemble regression. The system ingests consumer grade color camera mosaics of PV panels, classifies seven nuanced surface faults, and generates actionable severity grades for maintenance triage. By eliminating reliance on electroluminescence or IR sensors, our method enables affordable, scalable upkeep for resource limited installations, and advances the state of solar health monitoring toward universal field accessibility. Experiments on real public world datasets validate both classification and regression sub modules, achieving accuracy and interpretability competitive with specialized approaches.
\end{abstract}

%
%



\maketitle

\section{Introduction}
Photovoltaic energy is central to sustainable infrastructure, yet the physical durability of deployed modules is repeatedly challenged by weather, contaminants, and involuntary mechanical stress. Unaddressed, these superficial defects propagate and imperil overall plant yield. In the photovoltaics, operational constraints demand equally innovative yet resource frugal inspection frameworks. Acquisition of high fidelity visible-spectrum data is feasible via commodity imaging, but extracting reliable fault and severity information remains intellectually and practically challenging. Previous solutions incorporate sensor fusion, proprietary annotation formats, or cloud-based analytic workflows, which may be infeasible for remote or underserved locations. 
In this work, we build an integrated pipeline that leverages transformer networks for defect categorization and handcrafted features with ensemble regression for prioritization and predictive maintenance using a  standard color camera photo mosaics. 

\section{Related Work}
Surface defect analysis in PV modules has historically employed convolutional architectures, sometimes augmented by custom attention gating or transfer learning. However, these methods often focus on either detection or severity assessment in isolation, and may  require multi sensor aggregation \cite{Dwivedi} \cite{H Tella}. In recent years, transformer based networks have begun to outperform CNNs in image classification and localization due to their flexible inductive bias and superior feature aggregation; Its applications in solar inspection remain nascent. For severity grading, spatial-spectral features show promise for translating nuanced module damage patterns into ordinal or continuous scores aligned with technician judgment \cite{S. Ahmed} \cite{G. Cattani}. Our design merges advanced DL with classic ML for a modular, interpretable workflow tailored for decentralized field deployment.
\section{Methodology}
\subsection{Data Acquisition \& Preparation}
The dataset comprises planar visible spectrum images, acquired under diverse lighting and climatic scenarios from the public dataset. Each sample is labeled into Nine surface anomaly classes: Physical Damage, Bird Dropping, Clean, Electrical Fault, Snow Cover,Soiling,Cell damage,Breakage and Dust. Dataset augmentation uses controlled rotations 30 degrees, horizontal flips, and localized color jitter to simulate in-field variability, with all inputs resized to 224*224 pixels for model compatibility.  

\subsection{Defect Categorization using Vision Transformer}
A pretrained ViT-B/16 architecture, chosen for its balance of interpretability and feature expressiveness, is fine-tuned to Nine class output. Only the final classification head is updated to conserve compute and minimize overfitting given modest dataset scale. Input standardization and normalization promote transferability across varying acquisition devices.

\subsection{Severity Scoring using Regression Model}
Upon defect detection, segmented panel regions are analyzed for spectral and morpho metric attributes such as normalized area, edge density, color histogram entropy, and texture measures. These are input to a Random Forest regressor. The regression output is the severity grading (minor,major,nil), used for maintenance triage.

\begin{figure}[ht]
    \centering
    \includegraphics[height=10cm , width=0.43\textwidth]{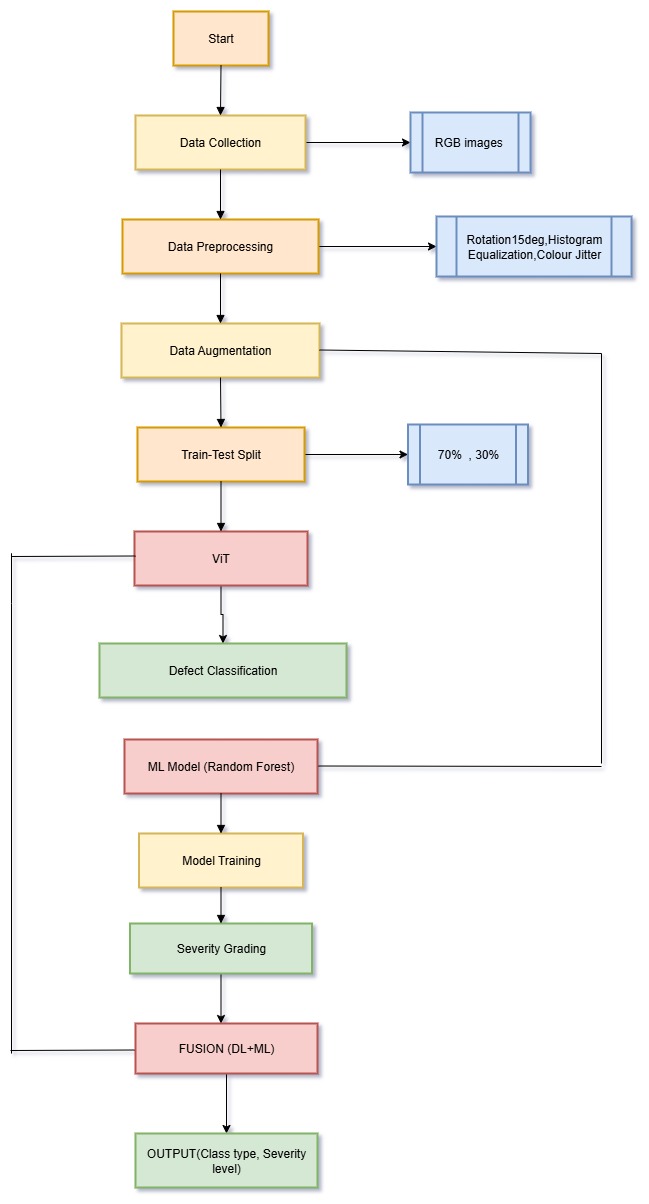}
    \caption{TinyViT Model Architecture}
    \label{fig:flowchart}
\end{figure}
\section{Experiments and Results}
\subsection{Dataset and Training Configuration}
The integrated pipeline was trained using 70\% of the panel mosaics (1,050 images), with the remaining 30\% (450 images) held for blind testing from a total sample of 1,500 panel images. All computation was performed on a single Nvidia T4 GPU for accessibility and reproducibility.

\subsection{Model Training and Optimization}
The transformer-based classifier was optimized using the AdamW optimizer with a learning rate of $3\times10^{-5}$, weight decay of $0.01$, and beta coefficients $(0.9, 0.999)$. To enhance model generalization, extensive data augmentation was applied including random resized cropping to $224\times224$ pixels with a scale range of $(0.7, 1.0)$, random horizontal and vertical flips, random rotation within $\pm30^\circ$, and color jitter transformations (brightness = 0.4, contrast = 0.4, saturation = 0.2, hue = 0.1). Regularization was implemented through dropout mechanisms with hidden layer dropout probability set to 0.3 and attention probability dropout configured at 0.3 to mitigate overfitting.

\subsection{Quantitative Performance}
The transformer-based classifier achieved an overall accuracy of 95\% and a macro F1-score of 94.7\% on defect categorization tasks. The ROC analysis demonstrated exceptional discriminative performance across all defect classes, with AUC values ranging from 0.90 (Physical-Damage) to 1.00 (Snow-Covered and Cell-Damage classes), indicating robust multi-class classification capability.

\begin{figure}[ht]
    \centering
    \includegraphics[width=0.43\textwidth]{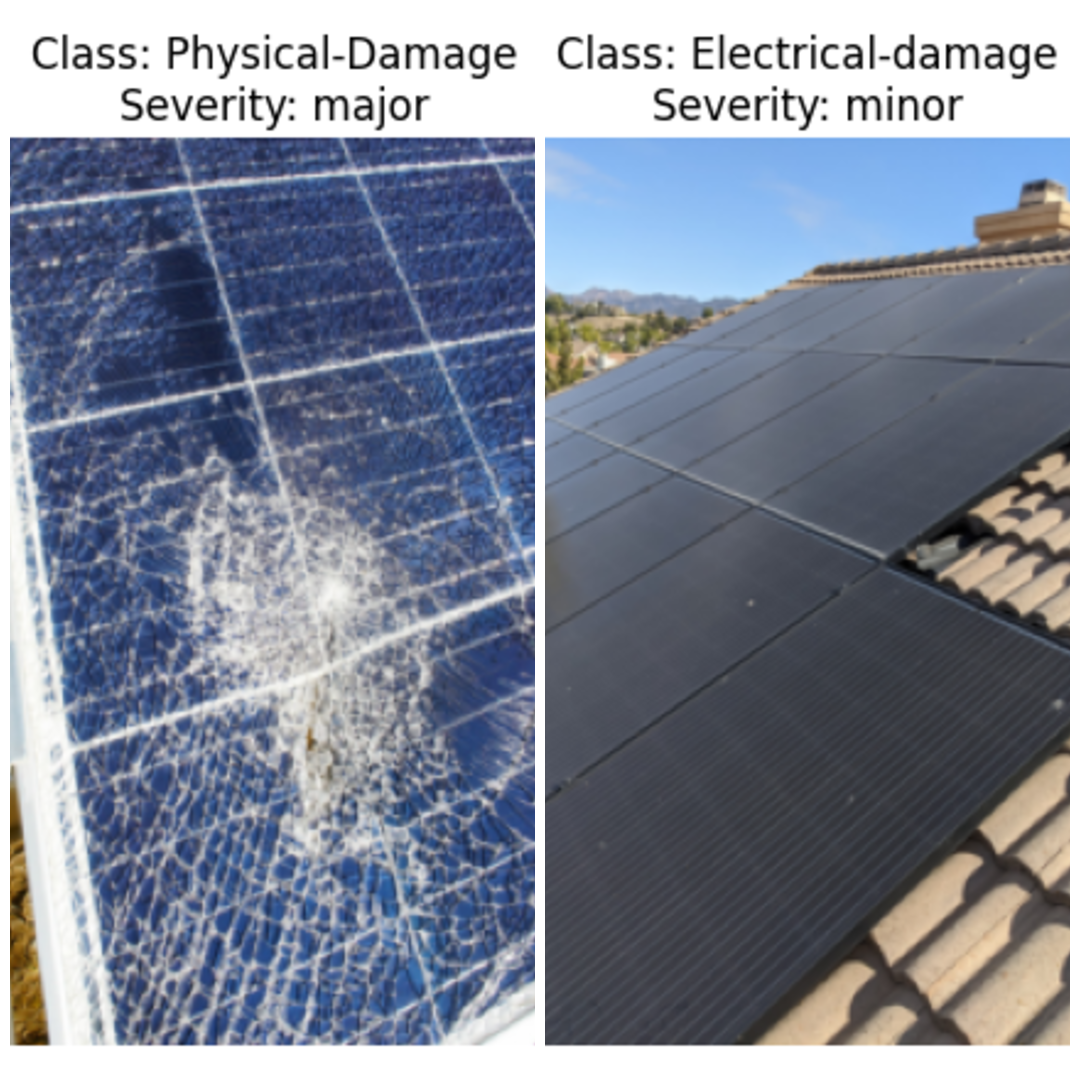}
    \caption{Fusion Model Final Output}
    \label{fig:photo-collage}
\end{figure}

\section{Conclusion}
This work demonstrates the effectiveness of customized transformer based and machine learning pipelines in extracting actionable health and severity insights from accessible, resource efficient visible-band panel images. By avoiding reliance on sensor fusion and computationally intensive approaches, TinyViT enhances the accessibility and scalability of solar plant maintenance solutions. Looking ahead, future directions include enabling seasonal defect identification, integrating performance data for holistic health monitoring, developing learning-based severity categorization, and achieving real-time deployment on edge hardware for decentralized, field-ready applications.
\begin{verbatim}

\end{verbatim}

\end{document}